\documentclass[conference,10pt,times,mathptm,psfig]{IEEEtran}
\usepackage{times}
\usepackage{epsfig}
\usepackage{graphicx}
\usepackage{amsmath}
\usepackage{amssymb}
\usepackage{pifont}
\usepackage{bm}
\usepackage{balance}
\usepackage{booktabs}
\usepackage{multirow}
\usepackage{threeparttable}
\usepackage{bbding}
\usepackage{hyperref}
\usepackage{cite}
\usepackage{algorithm}
\usepackage{algorithmic}
\usepackage{times}  
\usepackage{graphicx}  

\begin{document}
%
\title{T-SVG: Text-Driven Stereoscopic Video Generation}

\author{
Qiao Jin$^{1*}$\quad
Xiaodong Chen$^{1*}$\quad
Wu Liu$^{1\dag}$\quad
Tao Mei$^2$\quad
Yongdong Zhang$^1$\quad \\
$^1$University of Science and Technology of China, Hefei, China \quad $^2$Hidream Inc., Beijing, China \\
{\tt\footnotesize jq272@mail.ustc.edu.cn, cxd1230@mail.ustc.edu.cn, liuwu@ustc.edu.cn}
}

\maketitle

\begin{abstract}
The advent of stereoscopic videos has opened new horizons in multimedia, particularly in extended reality (XR) and virtual reality (VR) applications, where immersive content captivates audiences across various platforms. 
Despite its growing popularity, producing stereoscopic videos remains challenging due to the technical complexities involved in generating stereo parallax. 
This refers to the positional differences of objects viewed from two distinct perspectives and is crucial for creating depth perception. 
This complex process poses significant challenges for creators aiming to deliver convincing and engaging presentations.
To address these challenges, this paper introduces the Text-driven Stereoscopic Video Generation (T-SVG) system. 
This innovative, model-agnostic, zero-shot approach streamlines video generation by using text prompts to create reference videos. 
These videos are transformed into 3D point cloud sequences, which are rendered from two perspectives with subtle parallax differences, achieving a natural stereoscopic effect.
T-SVG represents a significant advancement in stereoscopic content creation by integrating state-of-the-art, training-free techniques in text-to-video generation, depth estimation, and video inpainting. 
Its flexible architecture ensures high efficiency and user-friendliness, allowing seamless updates with newer models without retraining. 
By simplifying the production pipeline, T-SVG makes stereoscopic video generation accessible to a broader audience, demonstrating its potential to revolutionize the field.

\end{abstract}


\IEEEpeerreviewmaketitle

\section{Introduction}
Stereoscopic videos have gained significant popularity in various applications, including entertainment, virtual reality (VR), and extended reality (XR) \cite{lavalle2023virtual, DBLP:books/daglib/0029461}, particularly with the emergence of devices such as Apple's Vision Pro and Meta's Quest. These devices have increased user interest in stereo content by making the stereo vision experience more accessible and convenient. However, the production of stereoscopic videos still faces significant challenges that hinder their widespread adoption.

Recent advancements in Artificial Intelligence Generated Content (AIGC) have transformed many aspects of media creation\cite{DBLP:conf/iclr/SingerPH00ZHYAG23, DBLP:journals/corr/abs-2311-15127, DBLP:conf/cvpr/ChenZCXWWS24}, especially in text to video generation models like Sora \cite{videoworldsimulators2024}. However, the techniques for generating stereoscopic videos have not fully capitalized on these innovations. Producing stereoscopic content presents unique challenges, particularly in generating stereo parallax, which involves the positional differences of objects viewed from two perspectives. Unlike the more accessible process of generating 2D videos, creating stereoscopic videos typically requires detailed 3D models or specialized stereoscopic recording equipment~\cite{DBLP:journals/ijcomsys/SuLKW11, 4379455, DBLP:books/daglib/0029461}. These traditional methods pose significant barriers and make the domain challenging for non-specialists.

\begin{figure}
  \includegraphics[width=0.95\linewidth]{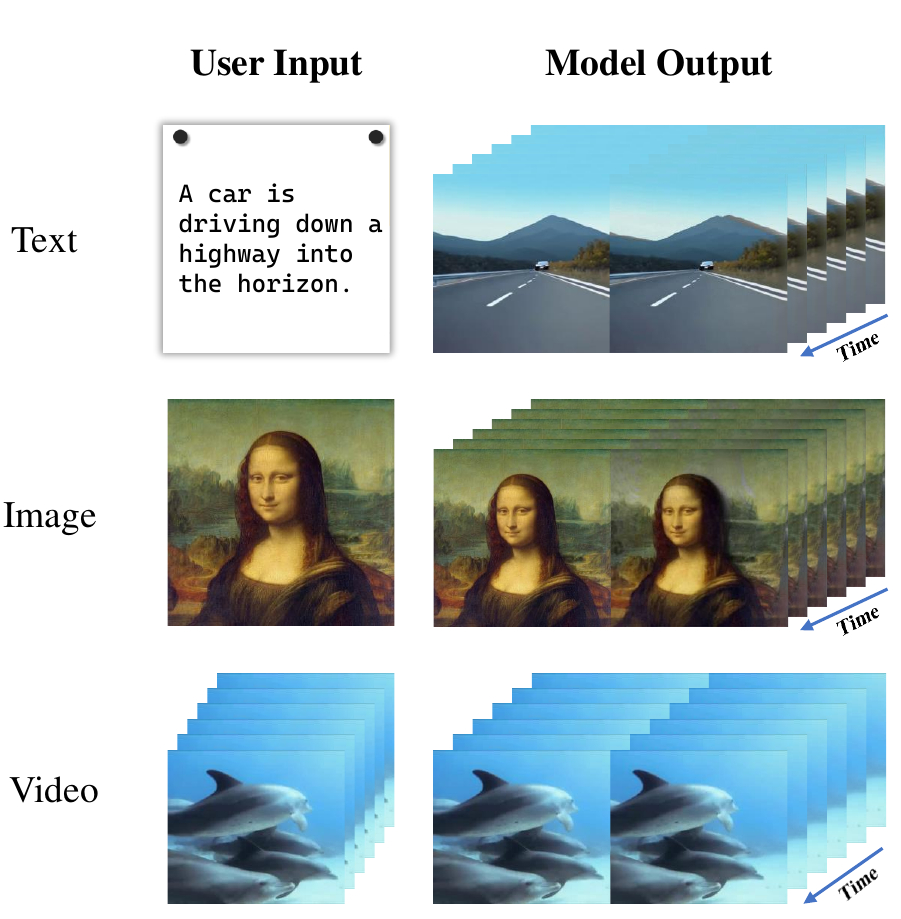}
  \caption{Illustration of T-SVG. T-SVG makes stereoscopic video generation accessible to broader users with texts, images and videos as prompt inputs}
  \label{fig:Demo}
\end{figure}

Given these challenges, we assert that the rapid development of deep learning models in computer vision has reached a maturity level where leveraging existing models, rather than training new ones for specific problems, can achieve similar results with greater computational efficiency. To this end, we introduce the Text-driven Stereoscopic Video Generation (T-SVG) system, an innovative, model-agnostic, zero-shot approach that optimizes the stereoscopic video generation process. By utilizing advanced methods in text-to-video generation, depth estimation, and video inpainting, T-SVG automates the creation of stereo pairs, ensuring high-quality output with minimal computational overhead. This system simplifies the production pipeline, making stereoscopic video generation more accessible to a wider audience.

\begin{figure*}
  \includegraphics[width=0.95\textwidth]{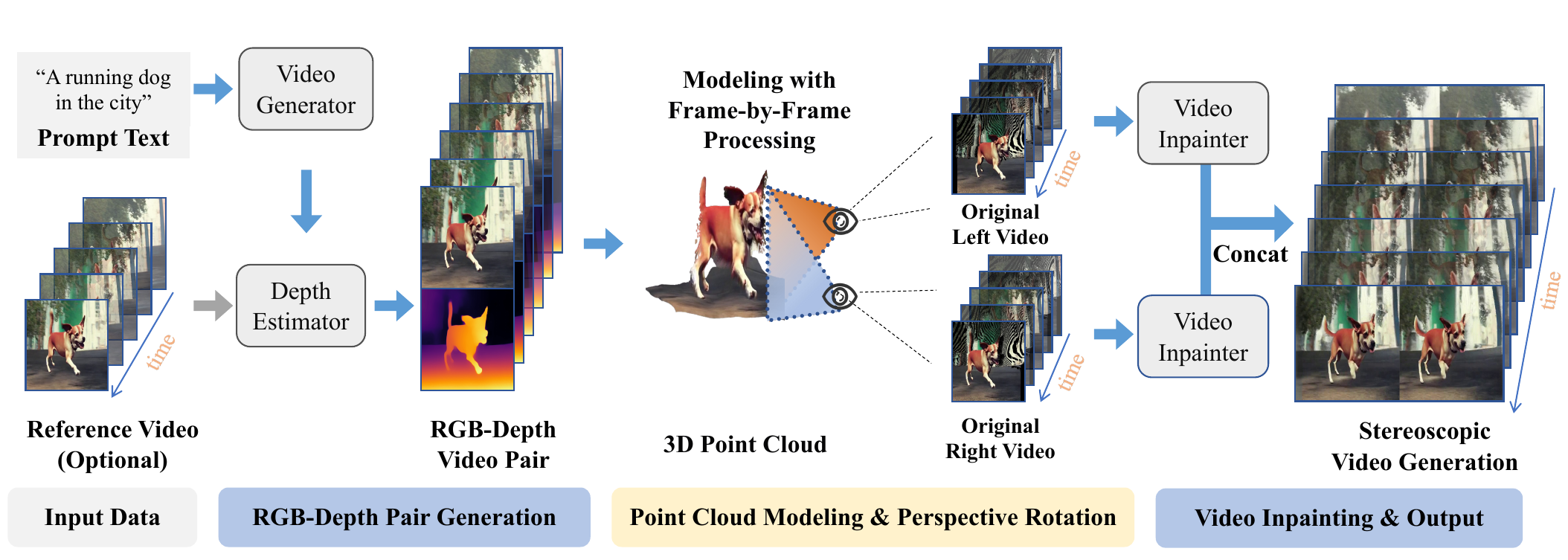}
  \caption{An overview structure of T-SVG. 1) First, we generate the video-depth pair according to the input text or video. 2) Then we generate 3D point clouds of each frame and apply perspective rotation of each side separately to get original stereoscopic videos. 3) At last, we use the video inpainting method to complete the original stereoscopic videos and concatenate them to get the stereoscopic video.}
  \label{fig:framework}
\end{figure*}

To achieve this, T-SVG begins with a textual prompt to generate a reference video, which is then processed to compute a depth map. This depth map is transformed into RGBD images and modeled as a 3D point cloud. Mimicking human binocular vision, these point clouds are rendered into a pair of videos with stereo parallax, producing a compelling stereoscopic visual effect.

T-SVG represents a significant advance in the creation of stereoscopic content by integrating state-of-the-art, training-free techniques in text-to-video generation, depth estimation, and video inpainting. The system’s flexible and modular architecture ensures high efficiency and user-friendliness, while also allowing for seamless updates with newer models without requiring retraining. This makes T-SVG highly adaptable to future advancements, offering substantial potential for evolving applications.

\section{Related Work}
In this paper, we propose a text-driven stereoscopic video generation system that integrates multiple deep learning models. This section reviews the development of text-to-video generation methods~\cite{cite:chen2022maple, cite:chen2024motion, cite:li2025m, cite:liu2024animateanywhere, cite:qu2024chatvtg} and their meaning to our method.

Text-to-video generation has evolved from text-to-image methods, aiming to create temporally coherent videos based on text input. To effectively capture the dynamics of video, temporal information is integrated into the original model through the incorporation of 2+1D convolutional layers or temporal transformer layers, extending from traditional image-based methods\cite{DBLP:journals/corr/abs-2210-02303, DBLP:conf/nips/HoSGC0F22, DBLP:conf/iscas/XuWLHZC24, DBLP:journals/corr/abs-2401-03048}. This allows for the extraction of video features, which are then controlled by text inputs through either GANs\cite{DBLP:journals/corr/GoodfellowPMXWOCB14} or diffusion-based\cite{DBLP:conf/nips/HoJA20, DBLP:conf/iclr/SongME21} architectures. In the beginning, GAN-based approaches often employed VAE\cite{DBLP:journals/corr/KingmaW13} or VQ-VAE\cite{DBLP:conf/nips/OordVK17} for video compression and reconstruction, gaining popularity due to their relatively low computational cost and high fidelity results. However, with the broader adoption of Latent Diffusion Models (LDM)\cite{DBLP:conf/cvpr/RombachBLEO22}, researchers have significantly reduced the computational burden of diffusion models by applying them in the feature space. This shift has allowed diffusion methods to surpass GANs as the primary approach for video generation, thanks to their more stable training processes.

The impressive results demonstrated by Sora have significantly surpassed those of previous models, validating the effectiveness of scaling laws\cite{DBLP:journals/corr/abs-2001-08361} in video generation. Current models, bolstered by increased computational power, are exploring new possibilities for generating longer videos, achieving higher resolutions, and enhancing controllability.  As the field of text-to-video generation advances rapidly, it provides a solid foundation for T-SVG, enabling it to effectively transform text inputs into videos and progress through subsequent production stages.

\section{System Architecture}
Our goal is to create an efficient text-driven system that generates stereoscopic videos from textual descriptions. As shown in Fig~\ref{fig:framework} and Algorithm~\ref{alg:stereo_video_gen}, our system comprises the following steps.

\begin{algorithm}[ht]
\caption{Text-Driven Stereoscopic Video Generation}
\label{alg:stereo_video_gen}
\begin{algorithmic}
\STATE \textbf{Input:} Text $T$
\STATE \textbf{Output:} Stereoscopic video $V_{stereo}$

\STATE $V \gets \text{GenerateVideo}(T)$
\STATE $D \gets \text{EstimateDepth}(V)$

\STATE Initialize $V_{L}, V_{R}$ as empty sequences

\FOR{each $(f_{rgb}, f_{D})$ in $(V, D)$}
    \FOR{each view $v$ in \{L, R\}}
        \STATE $ext_v \gets \text{GetExtrinsics}(P_{pcd}, v \text{ params})$
        \STATE $f_v \gets \text{RenderView}(P_{pcd}, ext_v)$
        \STATE Append $f_v$ to $V_{v}$
    \ENDFOR
\ENDFOR

\STATE $Mask_L, Mask_R \gets \text{GenerateMasks}(V_{L}, V_{R})$
\STATE $V_{L}, V_{R} \gets \text{Inpaint}(V_{L}, V_{R}, Mask_L, Mask_R)$

\STATE $V_{stereo} \gets \text{Combine}(V_{L}, V_{R})$
\RETURN $V_{stereo}$

\end{algorithmic}
\end{algorithm}

\subsection{Point Cloud Generation}

\begin{figure*}
  \includegraphics[width=0.95\linewidth]{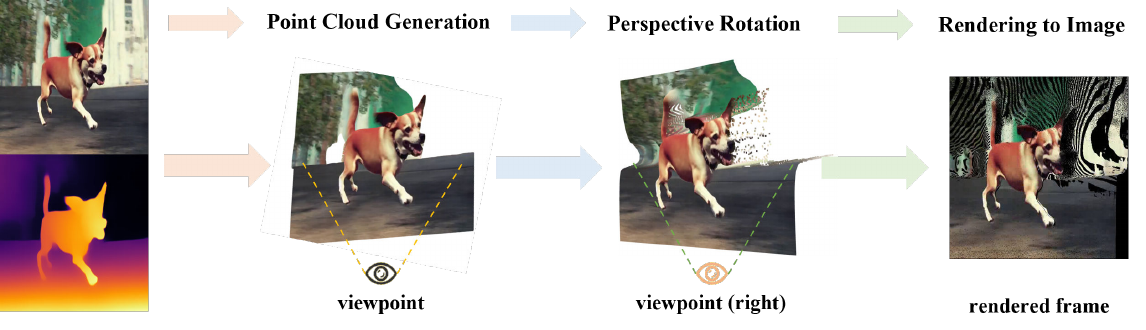}
  \caption{Details of Point Cloud Modeling and Perspective Rotation. The figure intuitively illustrates the processing steps and results of a single frame image in the Point Cloud Modeling and Perspective Rotation module. We first constructed a point cloud using the RGBD image; however, as shown in the figure, when the viewpoint is rotated, the points in the point cloud exhibit discontinuity and uneven distribution. When such a point cloud is rendered into 2D space, it leads to the phenomenon known as pixel dropout, where some pixels in the rendered image cannot find corresponding values.}
  \label{fig:figure4}
\end{figure*}

To produce stereoscopic videos, generating video pairs with subtle parallax disparities that mimic how each eye perceives depth is essential~\cite{Poggio1984}. A 3D model offers an optimal solution for simulating this disparity by placing two virtual cameras at positions corresponding to the human eyes, capturing the model from slightly different angles. To reduce computational complexity, we use a sparse point cloud representation instead of a highly detailed 3D model.

The process begins by extracting spatial information, such as depth, from the video. As illustrated in Fig.~\ref{alg:stereo_video_gen}, we first use a text-to-video generation model to create an initial video \( V \) based on a given text prompt \( T \). A depth estimator is then applied to generate depth maps for each frame, resulting in a set of RGB-D images that combine both color and depth data. From each RGB-D image, we generate a 3D point cloud through a series of mathematical transformations. For each pixel in the depth map \( D(x, y) \), its corresponding 3D point \( P(X, Y, Z) \) in the camera coordinate system is calculated using the following equations:

\begin{align}
X &= \frac{D(x, y) \cdot (x - c_x)}{f_x} \\[10pt]
Y &= \frac{D(x, y) \cdot (y - c_y)}{f_y} \\[10pt]
Z &= D(x, y)
\label{eq:1}
\end{align}

where \( f_x \) and \( f_y \) denote the camera's focal lengths, approximated to match the average focal length of the human eye. The parameters \( (c_x, c_y) \) refer to the image’s principal point, usually located at the center of the frame.

\subsection{Perspective Rotation}

Using the 3D point cloud model from the RGB-D image, we simulate the individual perspectives of each human eye for the stereoscopic video. This involves applying a transformation to mimic the natural parallax our eyes experience due to their horizontal separation. We transform the 3D point cloud to match a single eye's perspective using a composite transformation matrix. This matrix combines translation and rotation to replicate the essential natural parallax effect. The transformation matrix, denoted as \(\mathbf{M}\), is defined by the product of a rotation matrix \(R(\theta)\) around the Y-axis by an angle \(\theta\), and a translation matrix \(T\) along the X-axis by a distance \(t_x\). By applying this matrix to each point \(\mathbf{P}\) in the point cloud, we obtain the transformed coordinates \(\mathbf{P}'\):

\begin{equation}
\mathbf{M} = R(\theta) \cdot T =
\begin{bmatrix}
\cos(\theta) & 0 & \sin(\theta) & 1 \\
0 & 1 & 0 & 0 \\
-\sin(\theta) & 0 & \cos(\theta) & 0 \\
0 & 0 & 0 & 1
\end{bmatrix}
\cdot
\begin{bmatrix}
1 & 0 & 0 & t_x \\
0 & 1 & 0 & 0 \\
0 & 0 & 1 & 0 \\
0 & 0 & 0 & 1
\end{bmatrix}
\end{equation}

\begin{equation}
\vspace{10pt}
\mathbf{P}' = \mathbf{M} \cdot \mathbf{P} =\mathbf{M} \cdot
\begin{bmatrix}
X & Y & Z & 1
\end{bmatrix}^\top
\label{eq:3}
\end{equation}

To achieve the stereoscopic effect, transformed points are back-projected to generate distinct left and right views. However, due to the discrepancy between \( P' \) and \( P \) in Equation~\ref{eq:3}, back-projection may result in some pixels on the image plane lacking corresponding points in the transformed point cloud, a phenomenon termed 'pixel dropout'. As illustrated in~\ref{fig:figure4}, these pixels lack RGB values and are rendered as black. To mitigate this, we employ a video inpainting model that treats these regions as masks and fills them in, ensuring seamless and complete visual output in the final stereoscopic video.
\begin{table*}[ht]
\centering
\large
\caption{Model Parameters and Execution Times for the T-SVG System (Video Resolution: 480 × 540)} 
\label{tab:combined_stats}  
\renewcommand{\arraystretch}{1.1} 
\resizebox{0.98\textwidth}{!}{  
\begin{tabular}{l p{4.5cm} p{4.5cm} p{4.5cm}}  
\toprule
\textbf{Part} & \textbf{Model} & \textbf{Parameters (M)} & \textbf{Time (s)} \\ 
\midrule
\textbf{Video Generation} 
                     & Open Sora\cite{OpenSoraGitHub2024} & 5613.42 & 13.1344 \\ 
\midrule
\multirow{2}{*}{\textbf{Depth Estimation}} 
                     & DepthAnything\cite{depthanything} & 24.79  & 0.3405 \\ 
                     & NVDS\cite{nvds} & 93.22  & 1.1532 \\ 
\midrule
\multirow{1}{*}{\textbf{Stereo Frame Generation}} 
                     & - & - & 0.5420 \\ 
\midrule
\multirow{1}{*}{\textbf{Video Inpainting}} 
                     & ProPainter\cite{ProPainter} & 39.43  & 0.3667 \\ 
\midrule
\textbf{Total}     & - & \textbf{5613.42} & \textbf{15.5368} \\ 
\bottomrule
\end{tabular}
}
\normalsize 
\end{table*}

\begin{figure}
  \includegraphics[width=0.95\linewidth]{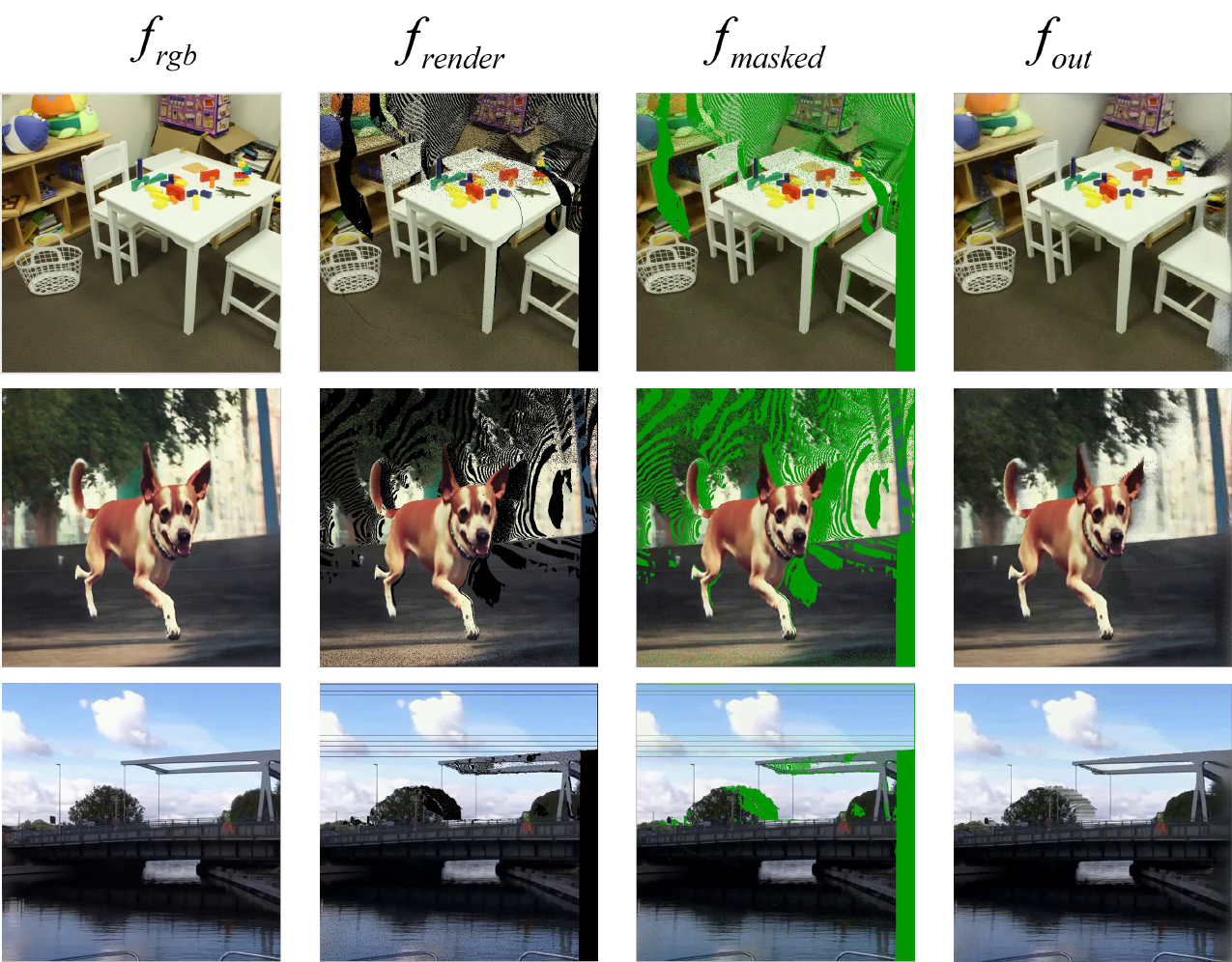}
  \caption{Illustration of pixel dropout and video inpainting. The first column shows a frame from different reference videos, and the second column illustrates the pixel dropout phenomenon in the images rendered from the point cloud after viewpoint switching, where some pixels are missing corresponding values. To repair these losses, T-SVG treats the areas with pixel dropout as mask regions that need to be filled in video inpainting, as shown in the third column. The fourth column presents the final inpainting result.}
  \vspace{-6mm}
  \label{fig:figure3}
\end{figure}

\section{Experiments and Result}

\subsection{Models Parameters and Time Cost}
As outlined in the System Architecture section and depicted in Figure~\ref{fig:framework}, our T-SVG system comprises three main components: Video-Depth Pair Generation, Point Cloud Generation, and Video Inpainting. The first and third components rely on existing deep learning models, all of which are interchangeable. Consequently, the selection of these models and the size of their parameters can significantly influence the performance of the generation system. In this section, we detail the specific models employed in each component, along with their respective parameter counts, as shown in Table~\ref{tab:combined_stats}. The testing was performed on an A40 GPU, with the video resolution set to 480 × 540 and the number of frames specified as 120 frames.

Table~\ref{tab:combined_stats} also presents the average per-frame execution time for each component within the system. As shown, the text-to-video generation process accounts for the majority of both parameter size and time consumption. In contrast, the time required by other parts remains within acceptable limits. Notably, the stereo frame generation section differs from other components in that it remains fixed within the system and operates independently of model changes, making its execution time particularly significant. The section is the key part of the T-SVG system, and the time cost of 0.5420 seconds per frame is relatively small. This allows border use of T-SVG and convenient applications in the future.

\subsection{Quantitative Results}
In this section, we examine the impact of each module on the final experimental results and present quantitative metrics and visualizations on the KITTI \cite{Menze2018JPRS} dataset.

We use four common metrics, including Peak Signal-to-Noise Ratio (PSNR), Structural Similarity Index (SSIM), Learned Perceptual Image Patch Similarity (LPIPS), and Fréchet Video Distance (FVD), to quantitatively assess the quality of generated videos. Together, these metrics evaluate fidelity, perceptual similarity, and motion consistency, offering a comprehensive measure of alignment between generated content and real-world video characteristics.

Table~\ref{tab:2} shows the quantitative results of our method and other methods. Since our approach is the first attempt in this field, we compare our method with the "leave blank" method, which only involves rendering without addressing pixel dropout. The results clearly show that our method performs better at all indexes.

\begin{table}[h]
\centering
\caption{Performance Comparison}
\renewcommand{\arraystretch}{1.2} 
\resizebox{0.48\textwidth}{!}{  
\begin{tabular}{@{}lcccc@{}}
\toprule
\textbf{Methods} & \textbf{PNSR↑} & \textbf{SSIM↑} & \textbf{LPIPS↓} & \textbf{FVD↓} \\
\midrule
Leave blank & 11.411 & 0.388 & 0.535 & 723.521\\
T-SVG & 12.793 & 0.474 & 0.411 & \textbf{398.563}\\
\bottomrule
\end{tabular}
}
\label{tab:2}
\end{table}

\section{Conclusion}
In conclusion, the Text-driven Stereoscopic Video Generation (T-SVG) technique marks a significant advancement in stereoscopic video creation. By utilizing text as the primary input, T-SVG simplifies the process and makes it accessible to a wider audience, including those with limited expertise in 3D content creation. The integration of cutting-edge methods in text-to-video generation, depth estimation, and video inpainting optimizes the production pipeline and enhances efficiency.

Importantly, the model-agnostic nature of T-SVG allows for the easy substitution of improved models as technology advances. This flexibility not only enhances the final results but also underscores the system's potential for continuous improvement. Ultimately, T-SVG opens new avenues for innovative applications in entertainment, education, and virtual experiences, transforming how stereoscopic content is produced and experienced.

\newpage
{
\balance
\bibliographystyle{IEEEtran}
\bibliography{./main}

\begin{thebibliography}{10}
\providecommand{\url}[1]{#1}
\csname url@samestyle\endcsname
\providecommand{\newblock}{\relax}
\providecommand{\bibinfo}[2]{#2}
\providecommand{\BIBentrySTDinterwordspacing}{\spaceskip=0pt\relax}
\providecommand{\BIBentryALTinterwordstretchfactor}{4}
\providecommand{\BIBentryALTinterwordspacing}{\spaceskip=\fontdimen2\font plus
\BIBentryALTinterwordstretchfactor\fontdimen3\font minus \fontdimen4\font\relax}
\providecommand{\BIBforeignlanguage}[2]{{%
\expandafter\ifx\csname l@#1\endcsname\relax
\typeout{** WARNING: IEEEtran.bst: No hyphenation pattern has been}%
\typeout{** loaded for the language `#1'. Using the pattern for}%
\typeout{** the default language instead.}%
\else
\language=\csname l@#1\endcsname
\fi
#2}}
\providecommand{\BIBdecl}{\relax}
\BIBdecl

\bibitem{lavalle2023virtual}
S.~M. LaValle, \emph{Virtual reality}.\hskip 1em plus 0.5em minus 0.4em\relax Cambridge university press, 2023.

\bibitem{DBLP:books/daglib/0029461}
\BIBentryALTinterwordspacing
T.~Matsuyama, S.~Nobuhara, T.~Takai, and T.~Tung, \emph{3D Video and Its Applications}.\hskip 1em plus 0.5em minus 0.4em\relax Springer, 2012. [Online]. Available: \url{https://doi.org/10.1007/978-1-4471-4120-4}
\BIBentrySTDinterwordspacing

\bibitem{DBLP:conf/iclr/SingerPH00ZHYAG23}
\BIBentryALTinterwordspacing
U.~Singer, A.~Polyak, T.~Hayes, X.~Yin, J.~An, S.~Zhang, Q.~Hu, H.~Yang, O.~Ashual, O.~Gafni, D.~Parikh, S.~Gupta, and Y.~Taigman, ``Make-a-video: Text-to-video generation without text-video data,'' in \emph{The Eleventh International Conference on Learning Representations, {ICLR} 2023, Kigali, Rwanda, May 1-5, 2023}.\hskip 1em plus 0.5em minus 0.4em\relax OpenReview.net, 2023. [Online]. Available: \url{https://openreview.net/forum?id=nJfylDvgzlq}
\BIBentrySTDinterwordspacing

\bibitem{DBLP:journals/corr/abs-2311-15127}
\BIBentryALTinterwordspacing
A.~Blattmann, T.~Dockhorn, S.~Kulal, D.~Mendelevitch, M.~Kilian, D.~Lorenz, Y.~Levi, Z.~English, V.~Voleti, A.~Letts, V.~Jampani, and R.~Rombach, ``Stable video diffusion: Scaling latent video diffusion models to large datasets,'' \emph{CoRR}, vol. abs/2311.15127, 2023. [Online]. Available: \url{https://doi.org/10.48550/arXiv.2311.15127}
\BIBentrySTDinterwordspacing

\bibitem{DBLP:conf/cvpr/ChenZCXWWS24}
\BIBentryALTinterwordspacing
H.~Chen, Y.~Zhang, X.~Cun, M.~Xia, X.~Wang, C.~Weng, and Y.~Shan, ``Videocrafter2: Overcoming data limitations for high-quality video diffusion models,'' in \emph{{IEEE/CVF} Conference on Computer Vision and Pattern Recognition, {CVPR} 2024, Seattle, WA, USA, June 16-22, 2024}.\hskip 1em plus 0.5em minus 0.4em\relax {IEEE}, 2024, pp. 7310--7320. [Online]. Available: \url{https://doi.org/10.1109/CVPR52733.2024.00698}
\BIBentrySTDinterwordspacing

\bibitem{videoworldsimulators2024}
\BIBentryALTinterwordspacing
T.~Brooks, B.~Peebles, C.~Holmes, W.~DePue, Y.~Guo, L.~Jing, D.~Schnurr, J.~Taylor, T.~Luhman, E.~Luhman, C.~Ng, R.~Wang, and A.~Ramesh, ``Video generation models as world simulators,'' 2024. [Online]. Available: \url{https://openai.com/research/video-generation-models-as-world-simulators}
\BIBentrySTDinterwordspacing

\bibitem{DBLP:journals/ijcomsys/SuLKW11}
\BIBentryALTinterwordspacing
G.~Su, Y.~Lai, A.~Kwasinski, and H.~Wang, ``3d video communications: Challenges and opportunities,'' \emph{Int. J. Commun. Syst.}, vol.~24, no.~10, pp. 1261--1281, 2011. [Online]. Available: \url{https://doi.org/10.1002/dac.1190}
\BIBentrySTDinterwordspacing

\bibitem{4379455}
S.~Knorr, A.~Smolic, and T.~Sikora, ``From 2d- to stereo- to multi-view video,'' in \emph{2007 3DTV Conference}, 2007, pp. 1--4.

\bibitem{cite:chen2022maple}
X.~Chen, W.~Liu, X.~Liu, Y.~Zhang, J.~Han, and T.~Mei, ``Maple: Masked pseudo-labeling autoencoder for semi-supervised point cloud action recognition,'' in \emph{Proceedings of the 30th ACM International Conference on Multimedia}, 2022, pp. 708--718.

\bibitem{cite:chen2024motion}
X.~Chen, W.~Liu, Q.~Bao, X.~Liu, Q.~Yang, R.~Dai, and T.~Mei, ``Motion capture from inertial and vision sensors,'' \emph{arXiv preprint arXiv:2407.16341}, 2024.

\bibitem{cite:li2025m}
A.~Li, X.~Chen, B.~Liang, Q.~Bao, and W.~Liu, ``M-adaptor: Text-driven whole-body human motion generation,'' in \emph{Proceedings of the Computer Vision and Pattern Recognition Conference}, 2025, pp. 2604--2613.

\bibitem{cite:liu2024animateanywhere}
H.~Liu, X.~Chen, X.~Liu, X.~Gu, and W.~Liu, ``Animateanywhere: Context-controllable human video generation with id-consistent one-shot learning,'' in \emph{Proceedings of the 5th International Workshop on Human-centric Multimedia Analysis}, 2024, pp. 41--43.

\bibitem{cite:qu2024chatvtg}
M.~Qu, X.~Chen, W.~Liu, A.~Li, and Y.~Zhao, ``Chatvtg: Video temporal grounding via chat with video dialogue large language models,'' in \emph{Proceedings of the IEEE/CVF Conference on Computer Vision and Pattern Recognition}, 2024, pp. 1847--1856.

\bibitem{DBLP:journals/corr/abs-2210-02303}
\BIBentryALTinterwordspacing
J.~Ho, W.~Chan, C.~Saharia, J.~Whang, R.~Gao, A.~A. Gritsenko, D.~P. Kingma, B.~Poole, M.~Norouzi, D.~J. Fleet, and T.~Salimans, ``Imagen video: High definition video generation with diffusion models,'' \emph{CoRR}, vol. abs/2210.02303, 2022. [Online]. Available: \url{https://doi.org/10.48550/arXiv.2210.02303}
\BIBentrySTDinterwordspacing

\bibitem{DBLP:conf/nips/HoSGC0F22}
\BIBentryALTinterwordspacing
J.~Ho, T.~Salimans, A.~A. Gritsenko, W.~Chan, M.~Norouzi, and D.~J. Fleet, ``Video diffusion models,'' in \emph{Advances in Neural Information Processing Systems 35: Annual Conference on Neural Information Processing Systems 2022, NeurIPS 2022, New Orleans, LA, USA, November 28 - December 9, 2022}, S.~Koyejo, S.~Mohamed, A.~Agarwal, D.~Belgrave, K.~Cho, and A.~Oh, Eds., 2022. [Online]. Available: \url{http://papers.nips.cc/paper\_files/paper/2022/hash/39235c56aef13fb05a6adc95eb9d8d66-Abstract-Conference.html}
\BIBentrySTDinterwordspacing

\bibitem{DBLP:conf/iscas/XuWLHZC24}
\BIBentryALTinterwordspacing
Z.~Xu, T.~Wang, D.~Liu, D.~Hu, H.~Zeng, and J.~Cao, ``Audio-visual cross-modal generation with multimodal variational generative model,'' in \emph{{IEEE} International Symposium on Circuits and Systems, {ISCAS} 2024, Singapore, May 19-22, 2024}.\hskip 1em plus 0.5em minus 0.4em\relax {IEEE}, 2024, pp. 1--5. [Online]. Available: \url{https://doi.org/10.1109/ISCAS58744.2024.10557902}
\BIBentrySTDinterwordspacing

\bibitem{DBLP:journals/corr/abs-2401-03048}
\BIBentryALTinterwordspacing
X.~Ma, Y.~Wang, G.~Jia, X.~Chen, Z.~Liu, Y.~Li, C.~Chen, and Y.~Qiao, ``Latte: Latent diffusion transformer for video generation,'' \emph{CoRR}, vol. abs/2401.03048, 2024. [Online]. Available: \url{https://doi.org/10.48550/arXiv.2401.03048}
\BIBentrySTDinterwordspacing

\bibitem{DBLP:journals/corr/GoodfellowPMXWOCB14}
\BIBentryALTinterwordspacing
I.~J. Goodfellow, J.~Pouget{-}Abadie, M.~Mirza, B.~Xu, D.~Warde{-}Farley, S.~Ozair, A.~C. Courville, and Y.~Bengio, ``Generative adversarial networks,'' \emph{CoRR}, vol. abs/1406.2661, 2014. [Online]. Available: \url{http://arxiv.org/abs/1406.2661}
\BIBentrySTDinterwordspacing

\bibitem{DBLP:conf/nips/HoJA20}
\BIBentryALTinterwordspacing
J.~Ho, A.~Jain, and P.~Abbeel, ``Denoising diffusion probabilistic models,'' in \emph{Advances in Neural Information Processing Systems 33: Annual Conference on Neural Information Processing Systems 2020, NeurIPS 2020, December 6-12, 2020, virtual}, H.~Larochelle, M.~Ranzato, R.~Hadsell, M.~Balcan, and H.~Lin, Eds., 2020. [Online]. Available: \url{https://proceedings.neurips.cc/paper/2020/hash/4c5bcfec8584af0d967f1ab10179ca4b-Abstract.html}
\BIBentrySTDinterwordspacing

\bibitem{DBLP:conf/iclr/SongME21}
\BIBentryALTinterwordspacing
J.~Song, C.~Meng, and S.~Ermon, ``Denoising diffusion implicit models,'' in \emph{9th International Conference on Learning Representations, {ICLR} 2021, Virtual Event, Austria, May 3-7, 2021}.\hskip 1em plus 0.5em minus 0.4em\relax OpenReview.net, 2021. [Online]. Available: \url{https://openreview.net/forum?id=St1giarCHLP}
\BIBentrySTDinterwordspacing

\bibitem{DBLP:journals/corr/KingmaW13}
\BIBentryALTinterwordspacing
D.~P. Kingma and M.~Welling, ``Auto-encoding variational bayes,'' in \emph{2nd International Conference on Learning Representations, {ICLR} 2014, Banff, AB, Canada, April 14-16, 2014, Conference Track Proceedings}, Y.~Bengio and Y.~LeCun, Eds., 2014. [Online]. Available: \url{http://arxiv.org/abs/1312.6114}
\BIBentrySTDinterwordspacing

\bibitem{DBLP:conf/nips/OordVK17}
\BIBentryALTinterwordspacing
A.~van~den Oord, O.~Vinyals, and K.~Kavukcuoglu, ``Neural discrete representation learning,'' in \emph{Advances in Neural Information Processing Systems 30: Annual Conference on Neural Information Processing Systems 2017, December 4-9, 2017, Long Beach, CA, {USA}}, I.~Guyon, U.~von Luxburg, S.~Bengio, H.~M. Wallach, R.~Fergus, S.~V.~N. Vishwanathan, and R.~Garnett, Eds., 2017, pp. 6306--6315. [Online]. Available: \url{https://proceedings.neurips.cc/paper/2017/hash/7a98af17e63a0ac09ce2e96d03992fbc-Abstract.html}
\BIBentrySTDinterwordspacing

\bibitem{DBLP:conf/cvpr/RombachBLEO22}
\BIBentryALTinterwordspacing
R.~Rombach, A.~Blattmann, D.~Lorenz, P.~Esser, and B.~Ommer, ``High-resolution image synthesis with latent diffusion models,'' in \emph{{IEEE/CVF} Conference on Computer Vision and Pattern Recognition, {CVPR} 2022, New Orleans, LA, USA, June 18-24, 2022}.\hskip 1em plus 0.5em minus 0.4em\relax {IEEE}, 2022, pp. 10\,674--10\,685. [Online]. Available: \url{https://doi.org/10.1109/CVPR52688.2022.01042}
\BIBentrySTDinterwordspacing

\bibitem{DBLP:journals/corr/abs-2001-08361}
\BIBentryALTinterwordspacing
J.~Kaplan, S.~McCandlish, T.~Henighan, T.~B. Brown, B.~Chess, R.~Child, S.~Gray, A.~Radford, J.~Wu, and D.~Amodei, ``Scaling laws for neural language models,'' \emph{CoRR}, vol. abs/2001.08361, 2020. [Online]. Available: \url{https://arxiv.org/abs/2001.08361}
\BIBentrySTDinterwordspacing

\bibitem{Poggio1984}
G.~F. Poggio and T.~Poggio, ``The analysis of stereopsis,'' \emph{Annual Review of Neuroscience}, vol.~7, no.~1, pp. 379--412, Mar. 1984.

\bibitem{OpenSoraGitHub2024}
\BIBentryALTinterwordspacing
hpcaitech, ``Open-sora: Democratizing efficient video production for all,'' GitHub, 2024, accessed: 2024/05/28. [Online]. Available: \url{https://github.com/hpcaitech/Open-Sora}
\BIBentrySTDinterwordspacing

\bibitem{depthanything}
L.~Yang, B.~Kang, Z.~Huang, X.~Xu, J.~Feng, and H.~Zhao, ``Depth anything: Unleashing the power of large-scale unlabeled data,'' \emph{CoRR}, vol. abs/2401.10891, 2024.

\bibitem{nvds}
Y.~Wang, M.~Shi, J.~Li, Z.~Huang, Z.~Cao, J.~Zhang, K.~Xian, and G.~Lin, ``Neural video depth stabilizer,'' in \emph{{ICCV}}.\hskip 1em plus 0.5em minus 0.4em\relax {IEEE}, 2023, pp. 9432--9442.

\bibitem{ProPainter}
S.~Zhou, C.~Li, K.~C.~K. Chan, and C.~C. Loy, ``Propainter: Improving propagation and transformer for video inpainting,'' in \emph{{ICCV}}.\hskip 1em plus 0.5em minus 0.4em\relax {IEEE}, 2023, pp. 10\,443--10\,452.

\bibitem{Menze2018JPRS}
M.~Menze, C.~Heipke, and A.~Geiger, ``Object scene flow,'' \emph{ISPRS Journal of Photogrammetry and Remote Sensing (JPRS)}, 2018.

\end{thebibliography}
}

\end{document}